\documentclass{article}
\usepackage{spconf,amsmath,graphicx}
\usepackage{url} 
\usepackage{wrapfig} 

\title{Local Statistics for Generative Image Detection}

\name{Yung Jer WONG and Teck Khim NG}
\address{National University of Singapore, School of Computing, Singapore}

\begin{document}
%
\maketitle
\begin{abstract}
Diffusion models (DMs) are generative models that learn to synthesize images from Gaussian noise. 
DMs can be trained to do a variety of tasks such as image generation and image super-resolution. 
Researchers have made significant improvements in the capability of synthesizing photorealistic images in the past few years. 
These successes also hasten the need to address the potential misuse of synthesized images. 
In this paper, we highlighted the effectiveness of Bayer pattern and local statistics in distinguishing digital camera images from DM-generated images. 
We further hypothesized that local statistics should be used to address the spatial non-stationarity problems in images. 
We showed that our approach produced promising results for distinguishing real images from synthesized images.  
This approach is also robust to various perturbations such as image resizing and JPEG compression.
\end{abstract}
\section{Introduction}
\label{sec:introduction}

In recent years, diffusion models, especially text-to-image diffusion models, have made synthetic image generation easy to consumers. 
As one of the most popular examples, Stable Diffusion (SD) \cite{rombach2022high} allows realistic and detailed images to be generated easily based on users' text prompts. 
In this paper, we focused on the problem of detecting real images from synthesized images. 
We designed features based on traces and patterns within real images that are not present in generated images. 
For the training of our classification models, we used \textbf{digital camera images} and  synthetic images. 
Even though synthetic images could be generated by many other types of Diffusion Models (DMs), we only used \textbf{SD-generated images} for training and showed that our models could also achieve excellent results even when tested on images generated by other types of DMs. 
We designed three localized feature sets and demonstrated why the combination of these feature sets could even survive JPEG compression and image resizing to successfully detect real digital camera images from DM-generated images. 
We presented the performance of our proposed methods on both uncompressed images and images that are resized and compressed.

\begin{figure}
  \centering
  \includegraphics[width=0.80\linewidth]{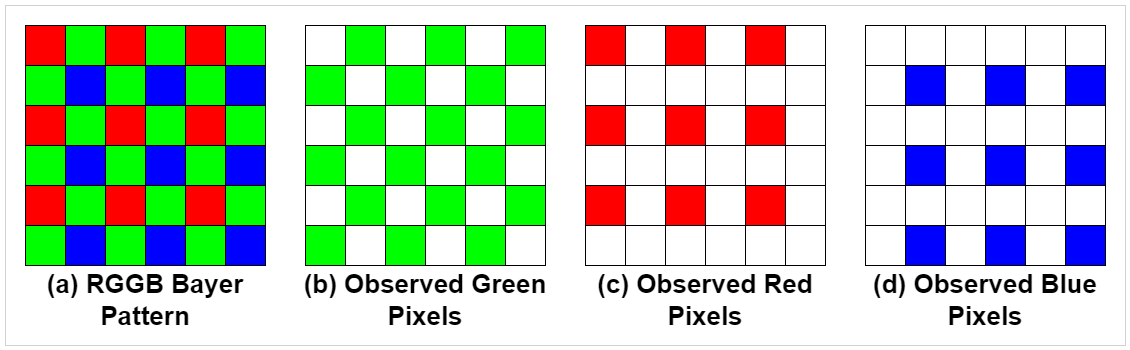}
  \caption{RGGB Bayer pattern in color filters. The green pixels are sampled twice as many as the red or blue pixels. Other possible Bayer patterns include BGGR, GBRG and GRBG.}
  \label{fig:bayerPattern}
\end{figure}

\section{Related Work}
\label{sec:related_work}

Existing methods for detecting real from fake images leveraged on handcrafted features such as texture analysis \cite{fu2022detecting}, image inconsistencies \cite{yang2019exposing}, co-occurrence features \cite{nataraj2019detecting} and artifact analysis \cite{li2018exposing}. 
In \cite{yu2019attributing} the authors substantiated the existence of artifacts in GAN-generated images. 
Subsequent research efforts highlighted the limitations of certain methods that were effective on GAN-generated images but not on DM-generated images. 
Corvi et al. \cite{corvi2023detection} showed that clear artifacts could be identified in the frequency domain of image residuals, extracted from the original images using the denoising filters. 
By leveraging deep CNNs and the Residual Neural Networks (ResNets) designed by He et al. in \cite{he2016deep}, the authors of \cite{corvi2023detection} managed to detect DM-generated images well. 
Wang et al. \cite{wang2023dire} utilized a pre-trained ADM \cite{dhariwal2021diffusion} as a reconstruction model and DDIM \cite{song2020denoising} process to compute the Diffusion Reconstruction Error (DIRE). 
A pre-trained ResNet-50 \cite{he2016deep} was then employed as the classifier. 
Most of the above mentioned methods involve the usage of deep neural networks to detect DM-generated images.


We noted that most digital cameras employ camera sensors with a Bayer filter, as shown in Figure \ref{fig:bayerPattern}. 
We observed a clear pattern: green pixels are sampled twice the number of red or blue pixels in the Bayer mosaic. 
The diagonals and anti-diagonals are either filled solely with green pixels or alternating with red and blue pixels. 
The digital camera's internal image processing pipeline implements a demosaicing algorithm to fully reconstruct three separate color channels. 
The values of the unobserved pixels in the three undersampled color channels are estimated with various interpolation methods such as bilinear and bicubic interpolation. 
This means that unobserved pixels are estimated during the reconstruction of the color channels.
Gallagher and Chen \cite{gallagher2008image} made use of this observation to detect digital camera images through frequency analysis on diagonal variances extracted from high-pass filtered images. 
They computed the variance of each diagonal from the green channel of any input image. 
For real images, these variances will exhibit an alternating pattern of high and low values that can be picked up using FFT. 
A peak will show up at half the normalized frequency. 
Synthetic images do not exhibit this property.
Gallagher and Chen's \cite{gallagher2008image} work however did not effectively address the problem of spatial non-stationarity. 
We addressed this and showed that our method is superior.

\nopagebreak[4]

\section{Methodology}
\label{sec:methodology}

In contrast to \cite{gallagher2008image}, we assumed that green pixels in Bayer mosaic are I.I.D. with variance only in a small local window. We proved below that interpolated pixels being weighted linear combinations of N neighboring pixels have smaller variance than the observed pixels: 

\begingroup
\setlength\abovedisplayskip{0pt}
  \begin{gather}
    B(x_i, y_i) = \sum_{(x_o, y_o) \in D} w_{(x_o, y_o)} B(x_o, y_o) 
  \end{gather}
\endgroup
\begingroup
\setlength\abovedisplayskip{0pt}
  \begin{align}
    \sum_{(x_o, y_o) \in D} w_{(x_o, y_o)} &= 1  \\
    var(B(x_o, y_o)) &= \sigma^2 \\
    var(B(x_i, y_i)) &= \sum_{(x_o, y_o) \in D} w_{(x_o, y_o)}^2\ \sigma^2 
  \end{align}
  \begin{gather*}
    0 < w_{(x_o, y_o)} < 1 \\
    0 < w_{(x_o, y_o)}^2 < w_{(x_o, y_o)} < 1\\
    0 < \sum_{(x_o, y_o) \in D} w_{(x_o, y_o)}^2 < \sum_{(x_o, y_o) \in D} w_{(x_o, y_o)} = 1\\
    var(B(x_i, y_i)) < var(B(x_o, y_o)) 
  \end{gather*}
\endgroup

\begin{figure}[tbp]
  \centering
  \includegraphics[width=0.85\linewidth]{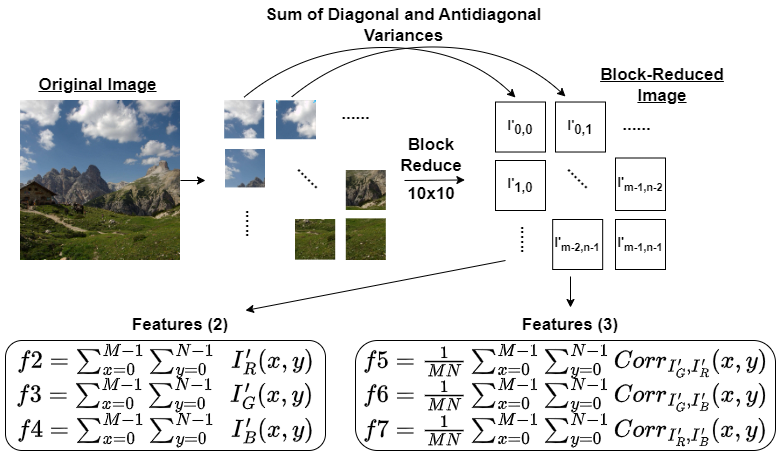}
  \caption{Flow diagram of the pipeline used to extract Features (2) and Features (3). \(I'\): Block-reduced Image. Each 10x10 block is reduced to one pixel representing the sum of diagonal and anti-diagonal variances. \((x,y)\): Current coordinates using a 0-based index. \(Corr_{A, B}\): Mean of local Pearson Correlation Coefficients between image A and B. \(M, N\): Number of pixels from the horizontal and vertical directions in \(I'\) respectively.
  }
  \label{fig:method23pipeline}
\end{figure}
where subscript $i$ represents``interpolated" pixels' coordinates; subscript $o$ represents ``observed" pixels' coordinates; $D$ is the set of all pixel coordinates that are used to estimate an unobserved pixel value; $B$ represents a local Bayer mosaic similar to the top-right diagram in Figure \ref{fig:bayerPattern}.  

We noted that the diagonals that span across various image regions suffer from spatial non-stationarity in pixel value statistics. 
Complex regions tend to have higher variance in pixel values due to the presence of sharp edges or rapid changes in pixel intensity. 
Flat regions tend to have lower variance in pixel values. 
We addressed this spatial non-stationarity problem in our proposed method. 
We designed three distinct feature sets computed using localized image processing techniques to mitigate the problems caused by spatial non-stationarity. 
To ensure naming consistency, we would refer them as \emph{Features (1)}, \emph{Features (2)} and \emph{Features (3)}. 

To compute Features (1), we first used a combination of a high-pass filter followed by a 2 $\times$ 2 filter that implements a localized gradient operation to pre-process the image. The 2 $\times$ 2 filters for diagonals and anti-diagonals were given respectively by the two kernels shown below: 
\begingroup
\setlength\abovedisplayskip{0pt}
\begin{table}[htbp]
  \centering
  \begin{minipage}{.12\textwidth}
    \centering
    \begin{equation}
      \begin{bmatrix}
        1 & 0 \\
        0 & -1 \\
      \end{bmatrix}
      \nonumber
    \end{equation}
  \end{minipage}%
  \begin{minipage}{.12\textwidth}
    \centering
    \begin{equation}
      \begin{bmatrix}
        0 & 1 \\
        -1 & 0 \\
      \end{bmatrix}
      \nonumber
    \end{equation}
  \end{minipage}
\end{table}
\endgroup

We computed the average of the absolute gradient for each diagonal and anti-diagonal from the high-pass filtered green channel. 
DFT was applied to the series of mean absolute gradients in the diagonal direction. 
The percentile at the normalized frequency of 0.5 compared to the other frequency components was noted. 
The same process was repeated for the anti-diagonal direction. 
These two percentiles form Feature (1). 
High percentiles are the evidence of Bayer Pattern interpolation.

When designing Features (2) and Features (3), we had to circumvent any potential artifact due to the 8x8 local block structure of JPEG compression. 
We empirically opted for a block size of 10 $\times$ 10 after conducting multiple experiments. 

For computing Features (2), we divided the image into non-overlapping blocks of 10 $\times$ 10 pixels.  
We computed the diagonal and anti-diagonal variances within each block. 
We then summed up all these variances i.e. each block was now represented by one number which is the sum of variances. 
The result was a \emph{block-reduced} image of variances which would also be used later in Features (3). 
Our last step for obtaining Features (2) was to sum up all the variances of the block-reduced image to obtain a single variance for each color channel. 


When designing Features (3), we noted that real digital camera images exhibit variations in pixel distribution among the color channels due to the differences in the number of sampled pixels in each channel. 
This will not be the case for synthetic images because diffusion models assume white noise that follows a Gaussian distribution. 
We defined Features (3) as the local Pearson correlation of the block-reduced image variances established in Features (2). 
For authentic images, the local correlation is lower than generative images due to the difference in numbers between green pixels and red or blue pixels sampled by the camera sensors. 
However, the local correlation for DM-generated images is higher because there is no demosaicing pattern that contributes to the difference in local correlations. 
All pixels in DM-generated images are the outputs of the backward diffusion process and image noise is assumed to follow a Gaussian distribution. 
We showed that block-reduced matrices of color channels are highly correlated due to the common assumption of additive noise following a Gaussian distribution.


In summary, we introduced a method consisting of three distinct feature sets. 
Each feature set utilized localized processing operations to effectively detect DM-generated images, even when they had been resized and compressed. 
We showed in our experiments that our method achieved state-of-the-art performance in detecting real from diffusion generated images. 
Figure \ref{fig:method23pipeline} shows the steps for extracting Features (2) and Features (3). 
In contrast to most of the recent works, our method is a non-deep learning approach that has advantages in that it is interpretable, generalizable, of low computational cost and requires very few training images.

\section{Experiment Setup}
\label{sec:experiment}
We conducted detailed analyses on sample images obtained from 2 sources: \emph{RAISE} dataset \cite{dang2015raise} and \emph{DiffusionDB} dataset \cite{wang2022diffusiondb}. 
We used the subset \emph{RAISE\_1k} from RAISE and the subset \emph{large\_random\_1k} from DiffusionDB with each subset consisting of \textbf{1000 images}.



We used \emph{RAISE} dataset as the only source of camera images because it is the only high quality camera image dataset publicly available. 
The images in \emph{RAISE} dataset were taken by different camera models which utilize the Bayer pattern. 

To show robustness and versatility, all experiment results of our proposed method were based on models that were trained using only  \textbf{10\%} (100 images) of the images from each dataset. 
The remaining \textbf{90 \%} (900 images) of each dataset served as the testing samples. 
We were able to use so few training images because our approach utilized deterministic features and it did not require a large number of training samples.

\section{Results and Discussion}
\label{sec:results_discussion}
In this section, we first showed the classification performance of our proposed method for randomly-cropped images of various sizes and also of various JPEG compression quality factors. 
We used image samples from \emph{RAISE} \cite{dang2015raise} and \emph{DiffusionDB} \cite{wang2022diffusiondb} datasets to represent digital camera images and DM-generated images respectively. 
The \emph{DiffusionDB} dataset was retrieved via HuggingFace \cite{lhoest2021datasets}.
The results were also compared with DIRE, one of the most recent generated-image detectors \cite{wang2023dire}, to demonstrate that our method is superior to DIRE in its generalizability in handling images from different diffusion models not included in training. 

\begin{figure}
  \centering
  \includegraphics[width=0.99\linewidth]{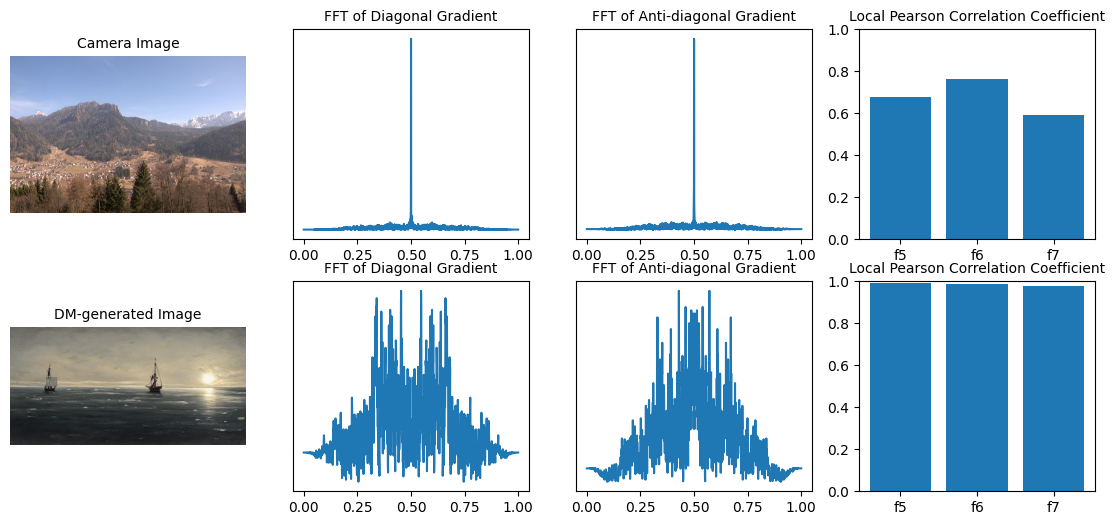}
  \caption{Visualization of Features (1) and Features (3). 
  Features (2) serve as a complement to Features (1) and Features (3). 
  \emph{Top}: A randomly sampled camera image (plot 1), the frequency analyses on diagonal gradients (plot 2) and antidiagonal gradients (plot 3) in the green channel, and a visualization of Features (3) for the camera image (plot 4). 
  \emph{Bottom}: A randomly sampled DM-generated image (plot 1), the frequency analyses on diagonal gradients (plot 2) and antidiagonal gradients (plot 3) in the green channel, and a visualization of Features (3) for the DM-generated image (plot 4). 
  }
  \label{fig:qualitative_comparison}
\end{figure}

Four distinct experiments were conducted to study the classification performance of our proposed method. 
In the $1^{st}$ experiment, our method was tested against randomly-cropped images of varying sizes.
We then fixed the dimensions of the images to 256 $\times$ 256 for the remaining three experiments. 
We conducted the experiments in the following manners: each image had \textbf{70\%} chance of undergoing JPEG compression ($2^{nd}$ experiment), image resizing ($3^{rd}$ experiment) or both ($4^{th}$ experiment). 
The image resizing procedures followed the IEEE VIP Cup 2022 standards \cite{corvi2023detection}\cite{rahman2023artifact}. 
Images which were not resized would be randomly-cropped to pre-determined dimensions.
The AUC scores and mean average precisions (mAP) of our method on all 4 experiments were displayed in Table \ref{table:tab1}.

\begin{table*}[htbp]
    \centering
    \resizebox{\textwidth}{!}{%
    \begin{tabular}{|c|c|c|c|c|c|c|c|c|c|c|c|c|c|c|}
      \hline
      \centering & \multicolumn{14}{c|}{\textbf{Our Method}} \\
      \hline
      \centering & \multicolumn{2}{c|}{\textbf{DiffusionDB}} & \multicolumn{2}{c|}{\textbf{ADM}} & \multicolumn{2}{c|}{\textbf{IDDPM}} & \multicolumn{2}{c|}{\textbf{PNDM}} & \multicolumn{2}{c|}{\textbf{DALL-E Mini}} & \multicolumn{2}{c|}{\textbf{GLIDE}} & \multicolumn{2}{c|}{\textbf{LDM}} \\
      \hline
      \centering\textbf{Size} & \textbf{Original} & \textbf{Resized} & \textbf{Original} & \textbf{Resized} & \textbf{Original} & \textbf{Resized} & \textbf{Original} & \textbf{Resized} & \textbf{Original} & \textbf{Resized} & \textbf{Original} & \textbf{Resized} & \textbf{Original} & \textbf{Resized}\\
      \hline
      1024 & 0.995/0.993 & 0.835/0.801 & \textbf{-}  & 0.813/0.758 & \textbf{-}  & 0.805/0.756 & \textbf{-}  & 0.824/0.773 & \textbf{-}  & 0.686/0.608 & \textbf{-} & 0.686/0.633 & \textbf{-}  & 0.787/0.735 \\
      512  & 0.960/0.950 & 0.777/0.729 & \textbf{-}  & 0.685/0.625 & \textbf{-}  & 0.684/0.619 & \textbf{-}  & 0.756/0.703 & \textbf{-}  & 0.697/0.662 & \textbf{-}  & 0.593/0.541 & \textbf{-}  & 0.744/0.701 \\
      256  & 0.975/0.967 & 0.821/0.808 & 0.974/0.962 & 0.765/0.698 & 0.977/0.965 & 0.757/0.673 & 0.983/0.974 & 0.783/0.720 & 0.980/0.973 & 0.802/0.795 & 0.971/0.959 & 0.695/0.652 & 0.985/0.978 & 0.836/0.829 \\
      128  & 0.918/0.904 & 0.802/0.800 & 0.939/0.926 & 0.704/0.666 & 0.941/0.928 & 0.671/0.615 & 0.953/0.943 & 0.738/0.699 & 0.948/0.933 & 0.810/0.813 & 0.921/0.901 & 0.683/0.666 & 0.953/0.941 & 0.845/0.842 \\
      64   & 0.902/0.888 & 0.771/0.775 & 0.912/0.900 & 0.663/0.633 & 0.922/0.910 & 0.658/0.621 & 0.923/0.913 & 0.692/0.671 & 0.911/0.896 & 0.743/0.758 & 0.899/0.883 & 0.694/0.684 & 0.933/0.922 & 0.774/0.776 \\
      \hline
    \end{tabular}}

    \vspace{0.05cm}

    \centering
    \resizebox{\textwidth}{!}{%
    \begin{tabular}{|c|c|c|c|c|c|c|c|c|c|c|c|c|c|c|}
      \hline
      \centering & \multicolumn{14}{c|}{\textbf{Our Method}} \\
      \hline
      \centering & \multicolumn{2}{c|}{\textbf{DiffusionDB}} & \multicolumn{2}{c|}{\textbf{ADM}} & \multicolumn{2}{c|}{\textbf{IDDPM}} & \multicolumn{2}{c|}{\textbf{PNDM}} & \multicolumn{2}{c|}{\textbf{DALL-E Mini}} & \multicolumn{2}{c|}{\textbf{GLIDE}} & \multicolumn{2}{c|}{\textbf{LDM}} \\
      \hline
      \centering\textbf{Quality} & \textbf{Compressed} & \textbf{RCRC} & \textbf{Compressed} & \textbf{RCRC} & \textbf{Compressed} & \textbf{RCRC} & \textbf{Compressed} & \textbf{RCRC} & \textbf{Compressed} & \textbf{RCRC} & \textbf{Compressed} & \textbf{RCRC} & \textbf{Compressed} & \textbf{RCRC}\\
      \hline
      95 & 0.903/0.880 & 0.847/0.852 & 0.941/0.918 & 0.639/0.603 & 0.944/0.918 & 0.631/0.582 & 0.955/0.933 & 0.661/0.627 & 0.951/0.927 & 0.824/0.826 & 0.871/0.856 & 0.654/0.641 & 0.963/0.939 & 0.839/0.837 \\
      90 & 0.909/0.887 & 0.839/0.829 & 0.929/0.909 & 0.663/0.617 & 0.940/0.917 & 0.652/0.592 & 0.953/0.929 & 0.708/0.671 & 0.919/0.878 & 0.831/0.825 & 0.853/0.841 & 0.642/0.628 & 0.946/0.908 & 0.862/0.847 \\
      80 & 0.907/0.871 & 0.850/0.845 & 0.934/0.902 & 0.595/0.569 & 0.936/0.899 & 0.578/0.540 & 0.951/0.923 & 0.630/0.600 & 0.921/0.880 & 0.808/0.807 & 0.868/0.840 & 0.618/0.612 & 0.947/0.904 & 0.820/0.812 \\
      60 & 0.905/0.881 & 0.842/0.833 & 0.927/0.903 & 0.617/0.592 & 0.927/0.899 & 0.595/0.556 & 0.951/0.929 & 0.676/0.636 & 0.933/0.902 & 0.822/0.797 & 0.851/0.828 & 0.641/0.614 & 0.961/0.944 & 0.846/0.827 \\
      40 & 0.912/0.902 & 0.848/0.848 & 0.929/0.902 & 0.647/0.610 & 0.927/0.894 & 0.644/0.581 & 0.947/0.923 & 0.709/0.667 & 0.934/0.923 & 0.823/0.818 & 0.833/0.821 & 0.622/0.600 & 0.962/0.943 & 0.861/0.855 \\
      20 & 0.921/0.905 & 0.850/0.837 & 0.927/0.904 & 0.655/0.637 & 0.922/0.890 & 0.651/0.591 & 0.944/0.920 & 0.698/0.673 & 0.937/0.922 & 0.815/0.815 & 0.856/0.833 & 0.644/0.621 & 0.961/0.946 & 0.856/0.847 \\
      \hline
    \end{tabular}}
  \caption{Cross-dataset evaluation for our method on DM-generated images from various sources to show generalizability. 
  ``\emph{RCRC}'' is the abbreviation for ``\textbf{R}andom-\textbf{C}ropping, followed by \textbf{R}esizing and \textbf{C}ompression''. 
  AUC/mAP were reported in each entry of the table. We substituted the SD-generated images from \emph{DiffusionDB} with DM-generated images from other sources. 
  Since generated images from various sources did not reach beyond the dimensions of 256 $\times$ 256, we ignored the results for 512 $\times$ 512 and 1024 $\times$ 1024 for original, uncompressed images. 
  ADM \cite{dhariwal2021diffusion}, IDDPM \cite{nichol2021improved}, PNDM \cite{liu2022pseudo} referred to the samples from their respective generative models sampled from the DiffusionForensics dataset \cite{wang2023dire}. 
  Dall-E Mini \cite{dayma2021DALLEMini}, GLIDE \cite{nichol2021glide} and LDM \cite{rombach2022high} referred to the samples from their respective generative models sampled from the generated image dataset used in \cite{corvi2023detection}. 
  The DM-generated images used to train our classifiers were solely from from \emph{DiffusionDB}. 
  All results were generated using 900 randomly selected images per class following our experimental settings.}
  \label{table:tab1}
\end{table*}

\begin{table}[htbp]
    \centering
    \resizebox{0.44\textwidth}{!}{%
    \begin{tabular}{|c|c|c|c|c|}
      \hline
      \centering & \multicolumn{2}{c|}{\textbf{Our Method}} & \multicolumn{2}{c|}{\textbf{DIRE \cite{wang2023dire}}}\\
      \hline
      \centering\textbf{Size} & \textbf{Original} & \textbf{Resized} & \textbf{Original} & \textbf{Resized}\\
      \hline
      1024 & 0.995/0.993 & 0.835/0.801 & 0.324/0.384 & 0.329/0.385 \\
      512  & 0.960/0.950 & 0.777/0.729 & 0.336/0.388 & 0.318/0.382 \\
      256  & 0.975/0.967 & 0.821/0.808 & 0.272/0.366 & 0.286/0.370 \\
      128  & 0.918/0.904 & 0.802/0.800 & 0.348/0.399 & 0.337/0.388 \\
      64   & 0.902/0.888 & 0.771/0.775 & 0.458/0.473 & 0.417/0.430 \\
      \hline
    \end{tabular}}
  \vspace{0.25cm}

    \resizebox{0.44\textwidth}{!}{%
    \begin{tabular}{|c|c|c|c|c|}
    \hline
    \centering & \multicolumn{2}{c|}{\textbf{Our Method}} & \multicolumn{2}{c|}{\textbf{DIRE \cite{wang2023dire}}}\\
    \hline
    \centering\textbf{Quality} & \textbf{Compressed} & \textbf{RCRC} & \textbf{Compressed} & \textbf{RCRC}\\
    \hline
    95 & 0.903/0.880 & 0.847/0.852 & 0.315/0.384 & 0.366/0.406 \\
    90 & 0.909/0.887 & 0.839/0.829 & 0.315/0.383 & 0.346/0.395 \\
    80 & 0.907/0.871 & 0.850/0.845 & 0.305/0.378 & 0.380/0.412 \\
    60 & 0.905/0.881 & 0.842/0.833 & 0.315/0.382 & 0.376/0.411 \\
    40 & 0.912/0.902 & 0.848/0.848 & 0.313/0.383 & 0.343/0.393 \\
    20 & 0.921/0.905 & 0.850/0.837 & 0.318/0.384 & 0.367/0.406 \\
    \hline
    \end{tabular}}

    \caption{Comparison between our method and the pre-trained DIRE detector \cite{wang2023dire} on the dataset shown in Table \ref{table:tab1}. 
    ``\emph{RCRC}'' is the abbreviation for ``\textbf{R}andom-\textbf{C}ropping, followed by \textbf{R}esizing and \textbf{C}ompression''. 
    We reported AUC and mAP respectively in each entry of the table (AUC/mAP). 
    This showed that deep-learning based detectors were not necessarily better than ordinary machine learning algorithms.
    \emph{Top}: The comparison of AUC and mAP scores between uncompressed images and images subjected to random-cropping followed by resizing across varying output sizes. 
    \emph{Bottom}: The comparison of AUC and mAP scores for images that were either only compressed or random-cropped followed by resizing and compression using various JPEG compression qualities. 
    Image outputs were standardized to dimensions of 256 $\times$ 256. 
    All the values were rounded to the nearest 3 s.f. }
    \label{table:tab2}
\end{table}

We included the AUC and mAP scores of the pre-trained DIRE model \cite{wang2023dire} in Table \ref{table:tab2} for comparison. 
Based on the results in Table \ref{table:tab1} and Table \ref{table:tab2}, our method produced convincing results in distinguishing real images from various DM-generated images. 
In contrast, the pre-trained DIRE model sufferred from performance degradation when applied to unseen real digital camera images and DM-generated images from different sources.
In addition to all the quantitative results shown above, we performed a qualitative comparison between two images randomly sampled from the \emph{RAISE} dataset and the \emph{DiffusionDB} dataset respectively by visualizing the aforementioned Features (1) and Features (3) as shown in Figure \ref{fig:qualitative_comparison}. Based on all the plots shown in the figure, we observed that camera images tend to have distinct peaks at half the normalized frequency in both diagonal and antidiagonal gradients in the green channel. The local Pearson correlation coefficients among the three color channels, as defined in Features (3), exhibited lower values in camera images compared to DM-generated images.

\section{Conclusion}
\label{sec:conclusion}
In this paper, we described a method combining three localized processing approaches and their corresponding feature sets to detect generative images. 
Instead of finding traces concealed in DM-generated images, we chose to find unique traces that are present in digital camera images but not in DM-generated images. 
Our approach was effective in distinguishing real images from DM-generated images.
We first demonstrated the excellent classification performance of our proposed method. 
We then compared the existing detector DIRE \cite{wang2023dire} with our method and showed that our method was superior than DIRE in terms of generalizability when handling unseen images. 
Qualitative comparison also showed significant differences between real and generated images when visualizing our proposed feature sets.
All experiment results showed that even though our method used only  \textbf{randomly selected 10\% of images for training}, it could still beat the performance of state-of-the-art deep-learning based classifiers. 


\bibliographystyle{IEEEbib}
\bibliography{strings,refs}

\end{document}